\documentclass[a4paper,12pt]{article} 
\usepackage{array}
\usepackage{adjustbox}
\usepackage{caption}
\usepackage{underscore}
\usepackage[a4paper,left=2.5cm,right=2.5cm,top=2.5cm,bottom=2.5cm]{geometry}
\usepackage{titlesec}
\usepackage{listings}
\usepackage{mathptmx}

\titleformat{\section}[hang]
{\normalfont\Large\bfseries}
{\thesection}{1em}{}
\usepackage[utf8]{inputenc}
\usepackage{amsmath}
\usepackage{graphicx}
\usepackage{hyperref}
\usepackage{float} 
\title{LLMs-in-the-Loop Part 2: Expert Small AI Models for Anonymization and De-identification of PHI Across Multiple Languages}
\author{Murat Gunay\footnote{Corresponding Author: murat.esra.gunay@gmail.com}, Bunyamin Keles\footnote{bkeles@mezun.hacettepe.edu.tr}, Raife Hizlan\footnote{raifehizlan@gmail.com}}
\begin{document}
\maketitle
\begin{abstract}
\noindent The rise of chronic diseases and pandemics like COVID-19 has emphasized the need for effective patient data processing while ensuring privacy through anonymization and de-identification of protected health information (PHI). Anonymized data facilitates research without compromising patient confidentiality.\\

\noindent This paper introduces expert small AI models developed using the LLM-in-the-loop methodology to meet the demand for domain-specific de-identification NER models. These models overcome the privacy risks associated with large language models (LLMs) used via APIs by eliminating the need to transmit or store sensitive data. More importantly, they consistently outperform LLMs in de-identification tasks, offering superior performance and reliability.\\

\noindent Our de-identification NER models, developed in eight languages — English, German, Italian, French, Romanian, Turkish, Spanish, and Arabic — achieved f1-micro score averages of 0.966, 0.975, 0.976, 0.970, 0.964, 0.974, 0.978, and 0.953 respectively. These results establish them as the most accurate healthcare anonymization solutions available, surpassing existing small models and even general-purpose LLMs such as GPT-4o.\\

\noindent While Part-1 of this series introduced the LLM-in-the-loop methodology for bio-medical document translation, this second paper showcases its success in developing cost-effective expert small NER models in de-identification tasks. Our findings lay the groundwork for future healthcare AI innovations, including biomedical entity and relation extraction, demonstrating the value of specialized models for domain-specific challenges.\\    
    
\noindent Keywords: de-identification, HIPAA, PHI, Patient Safety, LLMs-in-the-loop, Anonymization \end{abstract} 
\newpage

\section{Introduction}
Patient data is essential for improving public health, expanding preventive health services, preventing diseases, and formulating necessary health policies. Recent studies show that almost all (99\%) hospitals in the United States [1] use electronic health records (EHR). Similarly, Wales, Scotland, Denmark, and Sweden have adopted EHRs in the last few years. However, there is still a need for nationally accessible health data in the UK. In particular, the Covid-19 pandemic has again highlighted the importance of EHR data [2]. Thanks to EHRs, disease trends can be examined, modelling can be done, and health policies can be developed.\footnote{Our codebase will be available soon}.\\

Technology, which has become more complex and has developed with medical practices, necessitates the development of methods that will protect patient privacy [3].  With information security and information leakage recently gaining more importance, patient safety may have significant consequences beyond ethical violations in fundamental and health law [4].\\

Personal data is sensitive information that can be associated with an individual and is protected by various laws [5]. Personal privacy data in healthcare is called PHI and includes private information such as a patient's health history, treatments received, etc. [6].\\

EHRs contain both valuable clinical information and PHI. While EHRs are a rich data source for research, their usability is restricted due to the confidentiality of PHI [7-10]. For example, the HIPAA law regulates the use of 18 types of PHIs, such as name, phone number, dates, etc. (Table 1) [11, 12]. Therefore, PHI must be extracted from the text before EHR data can be used. Automating de-identification systems is needed since manually extracting PHIs is time-consuming and costly. In addition, coordination between annotators is also an important consideration [13, 14]. While early approaches to de-identification relied on complex rules to detect PHI, recent developments use machine learning methods and train on expert-annotated records. Hybrid systems integrate practices as features into statistical models, like conditional random fields (CRFs) [15].\\
\begin{table}[H]
    \caption{Protected Health Information Types [11]}    
    \centering
    \renewcommand{\arraystretch}{1} 
    \begin{tabular}{ll}
        \hline
        \textbf{No} & \textbf{PHI Type} \\ \hline
        1 & Names \\
        2 & All geographic subdivisions smaller than a state \\
        3 & Dates \\
        4 & Telephone numbers \\
        5 & Vehicle identifiers \\
        6 & Fax numbers \\
        7 & Device identifiers and serial numbers \\
        8 & Emails \\
        9 & URLs \\
        10 & Social security numbers \\
        11 & Medical record numbers \\
        12 & IP addresses \\
        13 & Biometric identifiers \\
        14 & Health plan beneficiary numbers \\
        15 & Full-face photographic images and any comparable images \\
        16 & Account numbers \\
        17 & Certificate/license numbers \\
        18 & Any other unique identifying number, characteristic, or code \\ \hline
    \end{tabular}    
    \label{tab:Table1}
\end{table}

The de-identification method makes it possible to use EHRs in research by removing confidential information [15]. Basic de-identification rules include removing direct identifying statements such as name, date, etc. Advanced statistical methods anonymize the data, thus reducing the risk of de-identification [16]. However, new techniques may also introduce unknown privacy risks. Therefore, continuous evaluation and improvement efforts are necessary [17]. Advanced methods can enable extensive collections of EHRs to be used efficiently and securely in research.\\

According to HIPAA, there are two possible methods of identity masking. The "Expert Determination" method, which requires employing an expert in the field, involves a small risk in identifying the individual whose information is used and is carried out using different statistical methods. In this method, the expert must have sufficient experience and knowledge. The other method is the "Safe Harbor" method, which involves de-identifying 18 pre-determined relevant identifiers that must be removed and/or modified from the corpus [11, 18]. In studies using deep learning (DL) models, the Safe Harbor method is used, and the relevant PHI is de-identified.\\

The lack of comprehensive data privacy frameworks can lead to vulnerabilities, leaving sensitive patient information susceptible to breaches and misuse. Despite efforts to anonymize this data, reidentification is still feasible through just a few spatiotemporal data points [19]. Recent advancements in privacy-preserving technologies have seen increased adoption [20], particularly in artificial intelligence (AI) and big data analytics. These technologies are vital in addressing major global health challenges by enhancing access to healthcare, promoting health, preventing diseases, and improving the overall experience for healthcare professionals and patients. AI, coupled with big data analytics, is the backbone for many innovations in digital health, driving improvements in care delivery and decision-making processes. Together, these domains are supported by additional technologies like the Internet of Things (IoT), next-generation networks (e.g., 5G), and privacy-preserving platforms such as blockchain [21, 22].\\

However, questions remain regarding accountability for AI and LLM outcomes. Since AI lacks autonomy and sentience, it cannot hold moral responsibility, leaving uncertainty about who should be accountable for its decisions and actions [23].\\

LLMs, particularly GPT-4, have demonstrated remarkable success in zero-shot de-identification tasks. However, significant challenges remain in utilizing proprietary paid APIs and open-source small LLMs. Paid APIs raise privacy concerns, as hospitals are often unwilling to transmit sensitive patient data to cloud-based services. Conversely, locally deployed small LLMs present difficulties in production due to their limited accuracy, resource-heavy requirements, and complex deployment processes. In response to these challenges, we propose Expert Small AI Models—lightweight, plug-and-play solutions that offer superior performance and accuracy compared to LLMs while being more practical and efficient for deployment in secure, on-premises environments. That’s why we don’t implement LLMs directly to solve this problem, we developed Expert Small NER Models with the LLM-in-the-loop methodology.\\
 
What is the LLM-in-the-loop?\\
 
From our perspective: the "LLM-in-the-loop" methodology LLMs is an integral part of the development process for expert small models, without relying on LLMs as the final solution. Instead of directly using LLMs for tasks, we utilize them selectively at various stages, such as synthetic data generation, rigorous evaluation, and agent orchestration, to improve the performance of smaller, domain-specific models. This approach allows us to benefit from the capabilities of LLMs while keeping the models efficient, focused, and specialized for specific tasks.\\

Recently, there has been a growing emphasis on the work done within the scope of LLM-in-loops. Studies have shown that LLMs perform better on tasks traditionally completed by humans [24-26], and the potential for effective utilization of LLMs is emphasized. It is seen that the innovative approach of "LLM-in-the-loop" is used in different fields today. In a study conducted to analyze social media content and reveal hidden themes [27]. In the study, the advanced capabilities of LLMs were leveraged to gain a deeper understanding of social media content by analyzing social media messages, discovering thematic structures and nuances in texts, and effectively matching texts to themes. Another study using the LLM-in-loops technique to improve the performance of LLMs was aimed at continuously improving the model outputs through iterative feedback loops, and this was applied in a study in the medical field. The aim was to increase the accuracy and reliability of the model and reduce hallucinations. The LLM-in-loops study, which involves the process of evaluating the model outputs as such human experts, giving feedback, and using this feedback to retrain the model, focused on reducing model errors and obtaining more reliable results in medical question-answer and summarization tasks [28].\\

Another study, which examines the potential of LLMs to recognize and examine intertextual relationships in biblical and Koine Greek texts, highlights how LLMs evaluate different intertextual scenarios and how these models can detect direct quotations, allusions, and echoes between texts. The study also mentions the ability of LLMs to generate intertextual observations and connections and the potential of these models to reveal new insights. However, it is noted that the model has difficulties with long query texts and can create incorrect intertextual connections, which reveals the importance of expert evaluation [29].\\

We first used the “LLMs-in-loop” method in the context of bio-medical document translation [30]. In this work, we demonstrate its success in developing cost-effective expert small NER models for de-identification tasks. Our findings lay the groundwork for future healthcare AI innovations, including biomedical entity and relation extraction, and demonstrate the value of specialized models for domain-specific challenges.\\

As we navigate the evolving landscape of AI in healthcare, the LLM-in-the-loop methodology stands out as a transformative approach. Recent studies have highlighted its capacity to enhance the performance of the models by leveraging human expertise to refine outputs continuously. This innovative strategy addresses the traditional challenges faced in biomedical text processing, such as accuracy and reliability, and mitigates issues like hallucinations that commonly occur in AI-generated content. By fostering a symbiotic relationship between human input and machine learning, we pave the way for advanced applications, including more effective biomedical entity extraction and improved medical summarization techniques. Ultimately, this research underscores the significance of integrating human intelligence with AI capabilities, setting the stage for more robust and trustworthy healthcare solutions.\\

\section{Background}
The de-identification model, called a Named Entity Recognition (NER) classification model, can be considered under four headings [31]:
\begin{itemize}
\item Rule-based models
\item Machine learning models
\item Hybrid models
\item Deep learning models
\end{itemize}
Techniques such as rule-based models and dictionaries can be easily implemented without labels but are vulnerable to input errors [31-34]. ML methods such as Support Vector Machines (SVM) and conditional random fields (CRF) can recognize complex patterns but require large amounts of labelled data and feature engineering and are poor at generalization [35-37]. Hybrid systems combine rule-based and ML models, providing high accuracy but requiring intensive feature engineering [38, 39].\\

Considering the disadvantages of the last three approaches to de-identification system creation, the latest state-of-the-art systems employ DL techniques to achieve better results than the best hybrid systems without requiring a time-consuming feature engineering process. DL is an ML subset using multilayered Artificial Neural Networks (ANN) and is very successful in most Natural Language Processing (NLP) tasks. Recent advances in DL and NLP (especially in the field of NER) enable the systems to outperform the winning hybrid system proposed by Yang and Garibaldi [39] on the 2014 i2b2 de-identification challenge dataset [31, 35].\\

De-identifying unstructured data is a widely recognized problem [40] in NLP, involving two key tasks: identifying PHI and replacing it through masking or obfuscation. Research has primarily focused on PHI identification. Early de-identification approaches [41] and  [42], especially in healthcare, were rule-based, using regular expressions, syntactic rules, and specialized dictionaries to detect PHI, such as phone numbers and emails. However, they struggled with identifying more complex entities like names and professions and required significant adjustments to function in different datasets, limiting their flexibility. The 2014 i2b2 project [34] introduced automatic de-identification, fueling the advancement of the machine and deep learning models for more accurate PHI detection. Early machine learning methods, such as Conditional Random Fields (CRF) [43], used hand-crafted features and lexical rules [44], signaling a shift to more adaptive and scalable approaches.\\

Work in the de-identification context has achieved human-level accuracy in de-identifying clinical notes from research datasets, but challenges remain in scaling this success to large, real-world environments. The hybrid context-based model outperformed traditional NER models by 10\% in the i2b2-2014 benchmark. It also has significantly fewer errors (93\% accuracy) compared to ChatGPT (60\% accuracy) [45].\\

LLM-based methods have been used in the development of de-identification models. However, these are still in the early stages, and further development is still needed to protect the privacy and security of health data [46]. The continued need to use APIs in LLM models and the problem of storing patient data reveals that expert models are still needed.\\

\section{Methodology} 
This section details the purpose of the research, the datasets employed, the methods for training and testing, the data preparation process, and the modelling and evaluation phases. Key to this study is the protection of personal data, adherence to legal regulations, and addressing the risks associated with processing sensitive patient information.\\

Our LLM-in-the-loop methodology leverages LLMs at key stages such as synthetic data generation, labelling, and evaluation, focusing on the development of high-performance, expert small models. To this end, we used a combination of proprietary closed-source data, open-source datasets, and synthetic data, all annotated by our labelling team in accordance with i2b2 labelling logic. The incorporation of synthetic data and LLM-assisted labelling further enhanced the scope and quality of our training datasets.\\

For English-language de-identification NER models, we utilized the entire dataset for training. The i2b2 test dataset served as the exclusive test set for evaluation purposes, allowing us to benchmark performance with high precision. For non-English languages, we applied an 80-20 split for training and testing. Additionally, our medical translation models [30] were used to translate the English datasets into non-English languages, generating high-quality parallel datasets across multiple languages.\\

In the data pre-processing phase, we employed language-specific tools to ensure accurate de-identification across different languages. The "Stanza" library was utilized for Romanian-language tasks, while the "NLTK" library was used for the other languages. Word tokenization for all datasets was performed using the "word-punct tokenizer" from the NLTK library.\\

For evaluation, we adopted the strict evaluation method, where both the chunk and the label had to match to be considered a correct prediction. This rigorous approach ensures the accuracy and reliability of our models, particularly in handling PHI.\\

By integrating proprietary, open-source, and LLM-synthesized datasets, as well as utilizing real and translated data, this methodology demonstrates the capability of expert small models to provide accurate, domain-specific de-identification solutions. Our approach minimizes reliance on large LLMs while ensuring privacy and top-tier performance in medical data anonymization.\\

The results in Table 4 and Table 7 were achieved using a structured and detailed prompt designed to extract Protected Health Information (PHI) from clinical notes. The prompt provided a comprehensive list of entity definitions, such as AGE, CITY, DEVICE, and ORGANIZATION, along with examples for clarity. It instructed GPT-4o to identify and mark entities using a consistent tagging format (e.g., BEGINER_LABEL CHUNK ENDNER) while preserving the original text. Specific guidelines were included for nuanced cases, such as excluding titles (e.g., "Dr.") from names and marking only actual dates for the DATE label. This rigorous approach ensured precision in high-performing categories and highlighted areas for improvement in more challenging entities. The prompt used in the study is presented in Appendix A.\\

\subsection{Datasets}
“i2b2-2014” is a research project \footnote{https://portal.dbmi.hms.harvard.edu} on de-identification and heart disease in clinical texts, and its labelling logic was used in our study. For English-language de-identification NER models, we utilized a combination of mostly open-source and synthetic data, with 22\% derived from proprietary closed-source data. The i2b2 test dataset served as the exclusive test set for evaluation, enabling us to benchmark performance with high precision. For non-English languages, we applied an 80-20 split for training and testing. Most of the non-English datasets were generated through translation from the English dataset using our medical translation models [30], open-source and through synthetic data generation with LLM-assisted labelling, producing high-quality parallel datasets across multiple languages.\\

Additionally, we utilized some NLP techniques and open-source third-party tools \footnote{LangTest by John Snow Labs: https://langtest.org/}  to enhance and augment the training datasets.\\

Although the i2b2 2014 dataset was not utilized for training purposes, we provide relevant information and statistics here to offer a more comprehensive understanding of its role in our evaluation process. i2b2/UTHealth is a dataset focused on identifying medical risk factors for Coronary Artery Disease (CAD) in the medical records of diabetic patients, where risk factors include hypertension, hyperlipidemia, obesity, smoking status, and family history, as well as diabetes, CAD, and indicators suggestive of the presence of these diseases [47]. i2b2 dataset consists of 1,304 progress notes of 296 diabetic patients. All PHIs in the dataset were removed throughout the study, and de-identification was performed randomly. The PHIs in this dataset were first categorized into HIPAA categories and then into i2b2-PHI categories, as shown in Table 2. 
Overall, the i2b2 dataset contains 56,348 sentences with 984,723 individual tokens, of which 41,355 are individual PHI tokens representing 28,867 particular PHI instances [31].\\

\begin{table}[H]
    \caption{PHI Categories, HIPPA, and Our Study Entities [31]}
    \centering
    \renewcommand{\arraystretch}{1} 
    \setlength{\tabcolsep}{5pt} 
    \begin{tabular}{p{3cm}p{5cm}p{5cm}}
        \hline
        \textbf{HIPPA} & \textbf{i2b2 Dataset} & \textbf{Our Dataset} \\ \hline
        Name & Patient, Doctor, Username & Patient, Doctor \\
        Profession & Profession & Profession \\
        Location & Street, City, State, Country, Zip, Hospital, Organization & Street, City, Country, Zip, Hospital, Location, Organization \\
        Age & Age & Age \\
        Date & Date & Date \\
        Contact & Phone, Fax, Email, URL, IP Address & Phone, Fax, Email \\
        ID & Medical Record, ID No, SSN, License No & Medical Record, ID, IDNUM, SSN \\
           &  & SEX \\
           &  & FAMILY \\ \hline
    \end{tabular}
    \label{tab:Table2}
\end{table}

In the literature review, it is seen that there are relative limitations in terms of data sets in de-identification model studies other than English. For this reason, it can be stated that only a few de-identification models have been developed for different languages. \\

In this respect, the de-identification models in different languages developed in this study will contribute to the literature and data scientists working on these models and the health institutions that will use them.\\

\subsection{Experimental Setup and Metrics}
\subsubsection{Clinical English de-identification Model}
The corpus of clinical admission discharge and private clinical reports from private hospitals and healthcare organizations was used to develop the English de-identification model. Labelling was done according to i2b2 2014 data principles as described previously. The labels used in the model, which uses a fine-tuned version of the "microsoft/deberta-v3-small" model as an embedding, are shown in Table 3.
\begin{table}[H]
    \caption{English De-identification Model Labels}
    \centering
    \renewcommand{\arraystretch}{1} 
    \setlength{\tabcolsep}{5pt} 
    \begin{tabular}{p{6cm}p{9cm}}
        \hline
        \textbf{Rule-Based Method Labels} & \textbf{Deep Learning Method Labels} \\ \hline
        ACCOUNT, DLN, EMAIL, FAX, IP, LICENSE, PLATE, SSN, URL, VIN & AGE, CITY, COUNTRY, DATE, DEVICE, DOCTOR, HOSPITAL, IDNUM, LOCATION-OTHER, MEDICAL RECORD, ORGANIZATION, PATIENT, PHONE, PROFESSION, STATE, STREET, USERNAME, ZIP \\ \hline
    \end{tabular}
    \label{tab:Table3}
\end{table}

In the study, ten labels were used for the Rule-based method, and 18 labels were used for deep learning methods. The training dataset was augmented for these labels since ORGANIZATION, PROFESSION, and LOCATION-OTHER entities gave low results due to the first training process with the deep learning method. The augmentation stages of the model were performed as follows:

\begin{itemize}
\item Firstly, a fake chunk data frame was created for each label in various formats. 
\item Sentences with the labels ORGANIZATION, PROFESSION, LOCATION-OTHER in the training dataset and CoNLL file were extracted.
\item Each labelled chunk was removed and replaced with label abbreviations.
\item The sentences were translated from English to the working language. For the translation, our medical translation models used [30].
\item The label abbreviations in the new sentences were replaced with new chunks of those labels from the fake data frame.
\item This new dataset was converted to “beginning, inside, outside” (BIO) format and added to the training dataset.\\
\end{itemize}
The model's performance implemented with the DL method used in this study was tested with the i2b2-2014 test set. It was observed that the retrained dataset with augmented labels showed better classification results when evaluated using the i2b2 2014 test set [33].\\

In the de-identification study conducted in English and with the DL method, learning rate=2e-5, max sentence length=512, batch size=2, and ten epoch train was performed. For the rule-based method, regexes suitable for each format were created for the selected labels. 
\\

\subsubsection{Non-English de-identification Models}
To understand which labels could be used in de-identification models and which labels would be appropriate for which aggregates and to determine the principles, the labelling team organized meetings with relevant hospital staff to make the models in German, French, Italian, Romanian, Spanish, and Turkish. Data collected from clinical admission reports, discharge reports, and special clinic reports obtained from hospitals and health institutions were labelled according to these principles.\\

The training process was carried out with the obtained data set. In the study, the 0.20 parts of the dataset determined during the division process were used as the test dataset. The dataset was preprocessed and converted into BIO format.\\

For German: bert-base-german-cased, for Italian: bert-base-italian-cased, for French: camembert-bio-base, for Romanian: bert-base-ro-cased, for Turkish: bert-base-turkish-cased and for Spanish: roberta-base-biomedical-clinical-es were used as embeddings.\\

The augmentation stages of the other language models were performed as follows:
\begin{itemize}
\item In the dataset used for the English in the de-identification model, a fake chunk data frame was created for each label in various formats. 
\item Each labelled chunk was removed and replaced with label abbreviations
\item The sentences were translated from English to the working language. For the translation, our medical translation models used [30].
\item The label abbreviations in the new sentences were replaced with new chunks of those labels from the fake data frame. 
\item This new data set was converted to BIO format and added to the train data set.
\end{itemize}

\noindent The de-identification research was performed with the DL method in seven languages other than English, learning rate=2e-5, max sentence length=512, batch size=16 (batch size=2 in Romanian), and ten epoch trains were performed.\\

\noindent\section{Result}
The results obtained from the de-identification NER models are shown in Table 4. In addition, the results obtained by using GPT-4o and the comparison results of other studies using the same dataset with the results obtained in this study are also included in the same table.
 
\begin{table}[H]
    \caption{De-identification Model in English i2b2-PHI Categories and Comparison (F1-Score)}
    \centering
    \renewcommand{\arraystretch}{1} 
    \setlength{\tabcolsep}{5pt} 
    \begin{tabular}{p{3cm}p{2cm}p{2cm}p{2cm}p{2cm}p{2cm}}
        \hline
        \textbf{PHI/Model Owners} & \textbf{Our Scores} & \textbf{Khin, Burckhardt [31]} & \textbf{Yang and Garibaldi [39]} & \textbf{Kocaman, Talby [45]} & \textbf{GPT-4o} \\ \hline
        AGE & 0.981 & 0.973 & 0.948 & 0.964 & 0.781 \\
        CITY & 0.944 & 0.909 & 0.776 & 0.949 & 0.917 \\
        COUNTRY & 0.881 & 0.805 & 0.303 & 0.920 & 0.802 \\
        DATE & 0.978 & 0.987 & 0.976 & 0.996 & 0.494 \\
        DEVICE & 0.762 & - & - & 0.286 & 0.217 \\
        DOCTOR & 0.966 & 0.962 & 0.945 & 0.980 & 0.743 \\
        HOSPITAL & 0.920 & 0.928 & 0.864 & 0.972 & 0.575 \\
        IDNUM & 0.867 & 0.756 & 0.838 & 0.909 & 0.288 \\
        LOCATION-OTHER & 1 & - & - & 0.722 & - \\
        MEDICAL RECORD & 0.942 & 0.979 & 0.971 & 0.980 & 0.716 \\
        ORGANIZATION & 0.876 & 0.719 & 0.427 & 0.874 & 0.400 \\
        PATIENT & 0.967 & 0.961 & 0.933 & 0.967 & 0.535 \\
        PHONE & 0.868 & 0.970 & 0.952 & 0.978 & 0.456 \\
        PROFESSION & 0.900 & 0.899 & 0.688 & 0.925 & 0.583 \\
        STATE & 0.961 & 0.932 & 0.863 & 0.969 & 0.932 \\
        STREET & 0.985 & 0.989 & 0.978 & 0.996 & 0.900 \\
        USERNAME & 0.962 & 0.957 & 0.978 & 0.954 & 0.635 \\
        ZIP & 0.989 & 0.982 & 0.986 & 0.982 & 0.975 \\ \hline
        Macro-avg & 0.931 & 0.919 & 0.840 & 0.863 & 0.548 \\ \hline
    \end{tabular}
    \label{tab:Table4}
\end{table}

\noindent As seen in Table 4, the model realized in this study includes PHIs not used in other studies, and satisfactory results were obtained. When the performance results of the studies are compared with the results of this study, it is determined that new SOTA values are obtained with this study. As a result of the analysis and calculations, although the train was performed with 18 PHI labels (DEVICE and LOCATION-OTHER labels were not used in other studies) and high scores of some labels were not obtained, the F1 macro score (0.931) obtained in this study was higher than the other models and a new SOTA value was received.\\

\noindent GPT-4o performs well in classes such as CITY, COUNTRY, ZIP, and STATE, achieving high precision, recall, and F1-scores. However, it struggles significantly with IDNUM, LOCATION-OTHER, ORGANIZATION, EMAIL, FAX, and DEVICE, where the scores are notably low. The macro average (0.5757) indicates that the model's performance varies significantly across classes, with weaker performance in certain categories. On the other hand, the micro average (0.5907) is slightly higher, reflecting the model's stronger performance in more frequent classes, but overall, the scores are low.\\

\noindent As a result of the de-identification study in seven different languages, the results obtained for 13 labels in German, Italian, and French are shown in Table 5, while the results obtained for Turkish (13 labels), Spanish (14 labels), and Romanian (14 labels) are shown in Table 6.\\

\begin{table}[H]
    \caption{German, Italian, and French de-identification Model Outputs (F1-Score)}    
    \label{tab:Table5}
    \centering
    \renewcommand{\arraystretch}{1} 
    \setlength{\tabcolsep}{5pt} 
    \begin{tabular}{lccc}
        \hline
        \textbf{Language/labels} & \textbf{German} & \textbf{Italian} & \textbf{French} \\ \hline
        AGE & 0.985 & 0.983 & 0.981 \\
        CITY & 0.963 & 0.922 & 0.939 \\
        COUNTRY & 0.954 & 0.906 & 0.926 \\
        DATE & 0.997 & 0.998 & 0.998 \\
        DOCTOR & 0.944 & 0.955 & 0.952 \\
        HOSPITAL & 0.981 & 0.975 & 0.915 \\
        IDNUM & 0.987 & 0.998 & 0.997 \\
        ORGANIZATION & 0.865 & 0.916 & 0.699 \\
        PATIENT & 0.903 & 0.920 & 0.918 \\
        PHONE & 0.995 & 0.995 & 0.995 \\
        PROFESSION & 0.980 & 0.917 & 0.941 \\
        STREET & 0.945 & 0.952 & 0.949 \\
        ZIP & 0.975 & 0.982 & 0.975 \\ \hline
        macro-avg & 0.960 & 0.955 & 0.937 \\ \hline
    \end{tabular}
\end{table}

The table presents F1-scores for de-identification tasks across German, Italian, and French datasets. Overall, the German model achieves the highest macro-average (0.960), followed by Italian (0.955) and French (0.937). DATE and PHONE categories exhibit consistently strong performance across all languages, achieving nearly perfect scores ($\geq$ 0.995). In contrast, the ORGANIZATION category shows notable variability, with the French model scoring significantly lower (0.699). These results highlight the robustness of the models in categories such as AGE, IDNUM, and ZIP while identifying areas for improvement in language-specific challenges, particularly for underperforming categories like ORGANIZATION in French (Table 5). However, since it was impossible to find any benchmark tests for these languages, comparing the scores obtained in this study was impossible.
 
\begin{table}[H]
    \caption{Turkish, Spanish, Romanian, and Arabic de-identification Model Outputs (F1-Score)}
    \centering
    \renewcommand{\arraystretch}{1} 
    \    \setlength{\tabcolsep}{4pt} 
    \begin{tabular}{>{\raggedright\arraybackslash}p{4.5cm}cccc} 
        \hline
        \textbf{\raggedright Language/labels} & \textbf{Turkish} & \textbf{Spanish} & \textbf{Romanian} & \textbf{Arabic} \\ \hline
        AGE & 0.988 & 0.980 & 0.984 & 0.980 \\
        CITY & 0.979 & 0.958 & 0.889 & 0.867 \\
        COUNTRY & 0.917 & 0.969 & 0.899 & 0.881 \\
        DATE & 0.997 & 0.997 & 0.973 & 0.987 \\
        DOCTOR & 0.953 & 0.969 & 0.966 & 0.908 \\
        EMAIL & - & 0.994 & 0.857 & - \\
        HOSPITAL & 0.942 & 0.976 & 0.935 & 0.988 \\
        ID & - & 0.995 & - & - \\
        IDNUM & 0.979 & - & 0.997 & 0.962 \\
        LOCATION & 1 & - & 0.846 & - \\
        MEDICAL RECORD & 1 & 0.991 & 0.999 & - \\
        ORGANIZATION & 0.975 & 0.734 & 0.768 & 0.978 \\
        PATIENT & 0.946 & 0.967 & 0.944 & 0.856 \\
        PHONE & 0.982 & 0.981 & 1 & 0.984 \\
        PROFESSION & 0.924 & 0.912 & 0.888 & 0.877 \\
        SEX & - & 0.971 & - & - \\
        SSN & - & 0.937 & - & - \\
        STREET & 0.913 & 0.959 & 0.953 & 0.768 \\
        ZIP & 0.913 & 0.980 & 0.992 & 0.950 \\
        FAX & - & - & 0.923 & - \\
        FAMILY & 1 & - & - & - \\ \hline
        macro-avg & 0.963 & 0.957 & 0.930 & 0.922 \\ \hline
    \end{tabular}
    \label{tab:Table6}
\end{table}

Table 6 highlights strong performances for Turkish (macro-avg 0.963) and Spanish (0.957) models, followed by Romanian (0.930) and Arabic (0.922). Categories like DATE, PHONE, and MEDICAL RECORD achieve near-perfect scores across languages, demonstrating model robustness. Lower scores are observed for CITY and ORGANIZATION in Romanian and Arabic, indicating room for improvement. Missing or language-specific labels (e.g., EMAIL, SSN) show variability in evaluation, reflecting dataset differences. Turkish and Spanish excel in most categories, with consistent performance across diverse labels.\\

\begin{table}[H]
    \caption{i2b2 Test Set Scores (IOB Token Level) Using GPT-4o}
    \centering
    \renewcommand{\arraystretch}{1} 
    \setlength{\tabcolsep}{6pt} 
    \begin{tabular}{lccc}
        \hline
        \textbf{Entity} & \textbf{Precision} & \textbf{Recall} & \textbf{F1-Score} \\
        \hline
        B-AGE & 0.688 & 0.937 & 0.791 \\
        B-CITY & 0.948 & 0.904 & 0.925 \\
        B-COUNTRY & 0.908 & 0.718 & 0.832 \\
        B-DATE & 0.808 & 0.834 & 0.821 \\
        B-DEVICE & 0.132 & 0.625 & 0.217 \\
        B-DOCTOR & 0.956 & 0.810 & 0.877 \\
        B-HOSPITAL & 0.916 & 0.675 & 0.775 \\
        B-IDNUM & 0.340 & 0.672 & 0.531 \\
        B-MEDICALRECORD & 0.960 & 0.794 & 0.869 \\
        B-ORGANIZATION & 0.303 & 0.695 & 0.422 \\
        B-PATIENT & 0.852 & 0.779 & 0.814 \\
        B-PHONE & 0.757 & 0.726 & 0.741 \\
        B-PROFESSION & 0.695 & 0.637 & 0.665 \\
        B-STATE & 0.902 & 0.974 & 0.937 \\
        B-STREET & 0.933 & 0.927 & 0.930 \\
        B-USERNAME & 0.563 & 0.728 & 0.635 \\
        B-ZIP & 1.000 & 0.993 & 0.997 \\
        I-AGE & 0.175 & 0.453 & 0.253 \\
        I-CITY & 0.872 & 0.852 & 0.862 \\
        I-COUNTRY & 0.800 & 0.615 & 0.696 \\
        I-DATE & 0.755 & 0.755 & 0.755 \\
        I-DEVICE & 0.133 & 1.000 & 0.235 \\
        I-DOCTOR & 0.490 & 0.767 & 0.605 \\
        I-HOSPITAL & 0.891 & 0.715 & 0.793 \\
        I-IDNUM & 0.392 & 0.550 & 0.458 \\
        I-LOCATION & 0.114 & 0.121 & 0.118 \\
        I-MEDICALRECORD & 0.763 & 0.457 & 0.571 \\
        I-ORGANIZATION & 0.246 & 0.750 & 0.370 \\
        I-PATIENT & 0.535 & 0.652 & 0.587 \\
        I-PHONE & 0.749 & 0.755 & 0.752 \\
        I-PROFESSION & 0.628 & 0.693 & 0.659 \\
        I-STATE & 0.917 & 0.688 & 0.786 \\
        I-STREET & 0.839 & 0.964 & 0.897 \\
        I-ZIP & 0.714 & 0.625 & 0.667 \\
        O & 0.986 & 0.984 & 0.985 \\
        \hline
        \textbf{Macro avg} & 0.5819 & 0.6247 & 0.5775 \\
        \textbf{Weighted avg} & 0.970 & 0.967 & 0.968 \\
        \hline
    \end{tabular}
    \label{tab:Table7}
\end{table}
Table 7 evaluates the B- (Beginning) and I- (Inside) tags separately, shows that the model achieves high accuracy (0.9672) overall. Classes like B-STATE, I-CITY, and I-COUNTRY perform very well, while B-EMAIL, B-FAX, and I-LOCATION have lower precision and recall, indicating challenges in identifying these entities. The macro average (0.5775) is lower than the weighted average (0.968), suggesting that less frequent or more difficult classes are pulling down the macro scores, whereas the model is quite successful in predicting the more common entities. \\

The Low Scores Are Attributed to Several Issues 
The model struggles to identify patient and doctor names located in the middle of the text, even though it can find those at the beginning and end. Some hospital names are partially labelled, which affects the overall precision and recall. Occasionally, the model includes extra tokens within the labels, leading to incorrect annotations. Despite specifying which labels to use in the prompt, the model sometimes incorrectly adds different labels, like time, which were not meant to be included. The model confuses some labels or fails to identify them altogether, contributing to the lower scores. 

\section{Conclusion}
This study underscores the importance of de-identification as a key method for safeguarding patient/personal health information and ensuring its ethical use in scientific research. By removing identifiable details through techniques like anonymization, generalization, and differential privacy, de-identification allows data to be used for diverse scientific applications, including epidemiological studies, disease modelling, and artificial intelligence development while maintaining patient privacy.\\

Recent advancements have demonstrated the potential of LLMs in de-identification tasks, yet challenges remain, particularly around issues of patient data security, API dependencies, and the need for domain-specific expertise in handling EHRs. Our "LLMs-in-the-loop" approach addresses these concerns by integrating small, specialized models tailored to the medical field. This method enhances both privacy and reliability, enabling the secure use of data without relying on external APIs or compromising sensitive patient information.\\

The multilingual nature of this research, spanning several languages, shows the adaptability and robustness of our models across diverse healthcare environments. While there are inherent risks associated with data anonymization, this study demonstrates that when properly applied, de-identification models can strike a delicate balance between protecting individual privacy and maximizing the utility of health data.\\

Furthermore, as the field progresses, it is crucial to establish globally recognized standards, raise awareness of best practices, and ensure that ethical principles guide the deployment of de-identification technologies. Transparency, accountability, and a rigorous risk-benefit analysis must remain at the forefront of these efforts.\\

Ultimately, the findings of this study highlight the potential of expert small models developed through the LLMs-in-the-loop methodology to meet the evolving demands of healthcare research. The models presented here offer a reliable and scalable solution for future de-identification applications, advancing the capabilities of AI in healthcare while safeguarding patient privacy.\\

Future research should focus on further refining and expanding de-identification models to cover a wider range of languages and healthcare contexts. One of the primary challenges is the scarcity of high-quality, annotated datasets in languages other than English, which limits the development of robust models for non-English speaking regions. Addressing this gap will require collaborative efforts to create and share multilingual datasets, ensuring more comprehensive language coverage. Additionally, future studies could explore more advanced augmentation techniques and develop models capable of handling increasingly complex medical data types, such as clinical narratives and imaging reports. Continuous innovation in privacy-preserving methods, such as federated learning, may also prove valuable in safeguarding sensitive patient information while advancing the performance and applicability of de-identification technologies across diverse healthcare systems.\\


\newpage 
Appendix A- The Prompt Used to Obtain Benchmarks with GPT-4o
\begin{lstlisting}
prompt = f"""You are tasked with extracting Protected Health Information (PHI) from clinical notes. Your job is to identify and mark specific entities within the text. Here are the entities you need to look for:

<entities>
AGE (Identifies the age number or age-related information. Example: In "88 years old," 88 would be marked as AGE. In "in his 50's,"50's would be marked as AGE.)
CITY (Identifies the name of a city.)
COUNTRY (Identifies the name of a country.)
DATE (Identifies specific dates or years. Example: In "He was admitted on 03/29/2089," 03/29/2089 would be marked as DATE. In "His surgery was in the 1980's," 1980's would be marked as DATE. In "His record was marked on 2089-08-24" 2089-08-24 would be marked at DATE.)
DEVICE (Identifies serial numbers, item code or product code of a medical device mentioned. Example: In "The AA 737 pacemaker was implanted," AA 737 would be marked as DEVICE.)
DOCTOR (Identifies the name of a doctor or healthcare professional. Only the name should be marked, not the title such as "Dr.", "M.D.".)
HOSPITAL (Identifies the name of a hospital or nursing home.)
IDNUM (Identifies identification numbers such as medical record or patient numbers.)
LOCATION (Identifies specific locations related to healthcare, excluding city or country.)
MEDICALRECORD (Identifies medical record numbers or similar identifiers.)
ORGANIZATION (Identifies names of organizations or institutions.)
PATIENT (Identifies the patient's name. Only the name should be marked, not titles like "Mr." or "Mrs.")
PHONE (Identifies phone numbers, including fax numbers.)
PROFESSION (Identifies professions or job titles.)
STATE (Identifies the name of a state or region.)
STREET (Identifies street addresses.)
USERNAME (Identifies usernames or account IDs.)
ZIP (Identifies postal or zip codes.)
</entities>

I will provide you with a clinical note. Your task is to process this note and mark all instances of the PHI entities listed above.

Here is the clinical note:

{clinical_note}

Instructions for marking PHI entities:
* Carefully read through the entire clinical note.
*  Identify any text that matches one of the PHI entity types listed above.
* For each identified PHI entity, mark the beginning and end of the relevant text chunk using the following format:
BEGINER_ LABEL CHUNK ENDNER where ENTITY LABEL is one of the entity types from the list, and CHUNK is the actual text containing the PHI.
* While marking, DO NOT EDIT OR CHANGE the original clinical text, only put marks described above.

Here are few examples of correct markup:

Original text:
Mrs. Linda Martinez, a 45-year-old architect, having MR\#: 2775283 for an evaluation on 2023-05-10. Her insulin pump model ZX900 was assessed by Dr. Michael Brown, M.D. The patient's condition has improved since the 1990s, but she mentioned feeling unwell for past 6 months. MF381/1183 was referenced during her visit, which lasted approximately 5 hours and concluded at 10:05:03. She was discharged on 20/10/2023.

Marked text:
Mrs. BEGINER_PATIENT Linda Martinez ENDNER, a BEGINER_AGE 45 ENDNER year-old BEGINER_PROFESSION architect ENDNER, having MR\#: BEGINER_MEDICALRECORD 2775283 ENDNER for an evaluation on BEGINER_DATE 2023-05-10 ENDNER. Her insulin pump model BEGINER_DEVICE ZX900 ENDNER was assessed by Dr. BEGINER_DOCTOR Michael Brown ENDNER, M.D. The patient's condition has improved since the BEGINER_DATE 1990s ENDNER, but she mentioned feeling unwell for past 6 months. BEGINER_IDNUM MF381/1183 ENDNER was referenced during her visit, which lasted approximately 5 hours and concluded at 10:05:03. She was discharged on BEGINER_DATE 20/10/2023 ENDNER.

Important notes:
* Be sure to process the entire clinical note and mark all instances of PHI entities.
* If a chunk of text could belong to multiple entity types, choose the most specific or appropriate one.
* Do not mark information that is not part of the specified PHI entity types.
* Preserve the original text exactly as it appears, including any spelling errors or formatting.
* Label the data, ensuring that professional titles or suffixes such as 'M.D.', 'Ph.D.', or similar are not removed. These titles must be preserved exactly as they appear in the text, without alteration or omission and should NEVER be inside the label.
* Apostrophe 's' ('s) should not be included within the label when associated with Names. Only the person's name should be inside the label, and the apostrophe 's' should remain outside the marked text. However, apostrophe 's' is allowed within the DATE label when referring to a decade (e.g., 80's).
* Mark only specific calendar dates as DATE. Do not mark relative time expressions like "6 months," "1 year ago," "5 weeks," "5 wks," "yesterday," "today," "days," or similar units of time (months, years, weeks), as they do not represent actual dates.
* Mark only actual dates as DATE. Do not mark time-related expressions such as "10:05:03," "10am," or durations like "5 hours" as DATE, since they refer to times or durations rather than specific calendar dates.
* Fax numbers should be treated as PHONE entities and marked the same way as phone numbers.
Please process the provided clinical note and return it with all PHI entities appropriately marked.
  """
\end{lstlisting}
\end{document}